\documentclass[sigconf]{acmart}
\usepackage{multirow} 
\usepackage{array}
\usepackage{makecell}
\usepackage{subcaption}
\usepackage{enumitem}

\AtBeginDocument{%
  \providecommand\BibTeX{{%
    \normalfont B\kern-0.5em{\scshape i\kern-0.25em b}\kern-0.8em\TeX}}}

\setcopyright{acmcopyright}
\copyrightyear{2018}
\acmYear{2018}
\acmDOI{10.1145/3627673.3679773}

\acmConference[Conference acronym 'XX]{}{June 03--05,
  2018}{Woodstock, NY}
\acmPrice{15.00}
\acmISBN{978-1-4503-XXXX-X/18/06}

\copyrightyear{2024}
\acmYear{2024}
\setcopyright{acmlicensed}\acmConference[CIKM '24]{Proceedings of the 33rd ACM International Conference on Information and Knowledge Management}{October 21--25, 2024}{Boise, ID, USA}
\acmBooktitle{Proceedings of the 33rd ACM International Conference on Information and Knowledge Management (CIKM '24), October 21--25, 2024, Boise, ID, USA}
\acmDOI{10.1145/3627673.3679773}
\acmISBN{979-8-4007-0436-9/24/10}
\begin{document}


\title{Do We Really Need Graph Convolution During Training? \\
Light Post-Training Graph-ODE for Efficient Recommendation}


\author{Weizhi Zhang}
\email{wzhan42@uic.edu}
\orcid{0000-0003-4067-7588}
\affiliation{%
  \institution{University of Illinois Chicago}
  \city{Chicago}
  \country{USA}}

\author{Liangwei Yang}
\email{lyang84@uic.edu}
\orcid{0000-0001-5660-766X}
\affiliation{%
  \institution{University of Illinois Chicago}
  \city{Chicago}
  \country{USA}}

\author{Zihe Song}
\email{zsong29@uic.edu}
\orcid{}
\affiliation{%
  \institution{University of Illinois Chicago}
  \city{Chicago}
  \country{USA}}

\author{Henry Peng Zou}
\email{pzou3@uic.edu}
\orcid{}
\affiliation{%
  \institution{University of Illinois Chicago}
  \city{Chicago}
  \country{USA}}

\author{Ke Xu}
\email{kxu25@uic.edu}
\orcid{0009-0001-2311-8090}
\author{Liancheng Fang}
\email{lfang87@uic.edu}
\orcid{}
\affiliation{%
  \institution{University of Illinois at Chicago}
  \city{Chicago}
  \country{USA}}

\author{Philip S. Yu}
\email{psyu@uic.edu}
\orcid{0000-0002-3491-5968}
\affiliation{%
  \institution{University of Illinois at Chicago}
  \city{Chicago}
  \country{USA}}


\begin{abstract}

The efficiency and scalability of graph convolution networks (GCNs) in training recommender systems (RecSys) have been persistent concerns, hindering their deployment in real-world applications. This paper presents a critical examination of the necessity of graph convolutions during the training phase and introduces an innovative alternative: the Light Post-Training Graph Ordinary-Differential-Equation (LightGODE). Our investigation reveals that the benefits of GCNs are more pronounced during testing rather than training. Motivated by this, LightGODE utilizes a novel post-training graph convolution method that bypasses the computation-intensive message passing of GCNs and employs a non-parametric continuous graph ordinary-differential-equation (ODE) to dynamically model node representations. This approach drastically reduces training time while achieving fine-grained post-training graph convolution to avoid the distortion of the original training embedding space, termed the embedding discrepancy issue. We validate our model across several real-world datasets of different scales, demonstrating that LightGODE not only outperforms GCN-based models in terms of efficiency and effectiveness but also significantly mitigates the embedding discrepancy commonly associated with deeper graph convolution layers. Our LightGODE challenges the prevailing paradigms in RecSys training and suggests re-evaluating the role of graph convolutions, potentially guiding future developments of efficient large-scale graph-based RecSys.
\end{abstract}





\maketitle

\section{Introduction}
Recommender systems (RecSys) are significant integral parts of many online platforms and web applications, helping users navigate vast amounts of information by providing personalized item recommendations. These systems are essential across various domains such as digital retailing ~\cite{wang2020time,hwangbo2018recommendation}, social networking platforms ~\cite{jamali2010matrix,fan2019graph}, and video-sharing services \cite{wei2023multi}, where they filter and tailor content to align with individual user preferences. Among the techniques \cite{pazzani2007content, van2013deep, koren2021advances, thorat2015survey} used in RecSys, collaborative filtering (CF) \cite{koren2021advances} is notably effective, and it predicts user preferences based on historical user-item interactions. Essentially, those historical interactions can be represented as a user-item bipartite graph. Inspired by the superior ability of graph convolution networks (GCNs) \cite{kipf2016semi, xu2018powerful, ma2023graph, wu2019simplifying} in modeling on graph-structured data, a large number of GCN-based recommendation models \cite{wang2019neural, wang2020disentangled, he2020lightgcn, ying2018graph} have emerged recently. They share the common idea of learning the node representation via acquiring neighborhood information in the bipartite graph layer by layer, thus capturing the multi-hop connectivity of users/items \cite{wu2022graph}. 

Despite the inspiring progress made in graph-based recommendation, these approaches are inherently challenged by the issues of efficiency and scalability. They are intrinsically raised by the computation-intense message-passing of graph convolution in the existing training paradigm of graph-based recommendation. Such problems are further exaggerated in the real-world application of large-scale graphs as the time/computation complexity will grow exponentially with the number of users and items. 
Recent studies show that simple MLPs as the initialization of graph model \cite{yang2023graph, han2022mlpinit} or trained with contrastive learning \cite{hu2021graph}, knowledge distillation \cite{zhang2021graph} demonstrate competitive performance compared with GCN models as long as they share an equivalent weight space. Considering that one can trivially derive a counterpart light graph model \cite{he2020lightgcn} given the matrix factorization (MF) \cite{mnih2007probabilistic} weight, we naturally raise a meaningful and significant question:  \textit{Do we really need computation-intense graph convolution during training for recommendation?}

To address the inquiry, we first conducted a preliminary experiment to investigate the role of graph convolution. 
The results reveal that graph convolution has a more pivotal role in testing rather than in training. Notably, the MF model is capable of matching the performance of the GCN when a similar lightweight graph convolution \cite{he2020lightgcn} is implemented after training. To uncover the underlying reasons from a training viewpoint, we examined the supervision alignment force when training with MF and LightGCN models, finding that the alignment property \cite{wang2020understanding, wang2022towards} of positive user-item pairs is approximate in two distinct training paradigms. This prompted us to further explore the training processes of the MF and GCN models, leading us to conclude that GCN-based training essentially acts as a degree-weighted form of MF training. Intuitively, by following the pairwise alignment force from a depth-first search (DFS) perspective, MF training results in effects akin to GCN training that adopts information aggregation based on breadth-first search (BFS). Given the time demand of these processes, we suggest that graph convolution may not be necessary during training. 
However, the current graph convolution method is suboptimal, as we empirically find that the increasing number of layers significantly enlarges the difference between embeddings before and after convolution, denoted as the \textbf{Embedding Discrepancy}.
Assuming the MF model is well-trained, any post-training operations should not significantly alter the original embedding space, whereas the existing convolution strategy with high embedding discrepancy may potentially offset the benefits of higher-order information. Moreover, the existing coarse-grained graph convolution approaches fail to find an optimal convolution depth due to its discrete characteristics. These motivate us to seek a more fine-grained method to integrate higher-order user-item interactions while avoiding computation-intense message passing during training.

In this paper, we introduce Light Post-Training Graph-ODE (LightGODE), a novel graph-based method designed for fine-grained and efficient large-scale RecSys. Specifically, we first propose a novel Post-Training Graph Convolution (PTGC) paradigm that significantly improves training efficiency by skipping the most time-consuming operations, including adjacency matrix normalization and layer-by-layer graph convolutions, making the training process as efficient as for traditional MF models. To tackle the issue of embedding discrepancy, we develop a non-parametric graph convolution that incorporates the self-loop during the information update. This straightforward operation will prioritize the preceding layers, thereby implicitly assigning greater importance to shallow layers, particularly the initial embeddings during graph convolution, which aids in minimizing the variation between the embedding spaces before and after graph convolution. Thus, it helps to reduce the distribution discrepancy between the embedding space. Building on this foundation, we propose a continuous graph ordinary-differential-equation derived from the discrete parameter-free graph convolution. The continuity offers several advantages. First, it characterizes the continuous dynamics of user/item representations within the bipartite graph, making the traditional graph convolution a specific discretization of seamless layer-wise embedding transformation. Additionally, it enables precise and fine-grained graph convolution to achieve the optimal trade-off with the continuous-time value to capture high-order information while balancing the embedding discrepancy. 
To foster future research and development of LightGODE, we have released it open-source, available at \textcolor{blue}{\url{https://github.com/DavidZWZ/LightGODE}}. Here, we summarize our contributions as follows:

\begin{itemize}[leftmargin=*]
    \item To the best of our knowledge, we are the first to challenge the long-standing authority in the graph-based recommendation - the necessity of graph convolution, and we empirically and analytically reveal its decisive role in testing rather than training.
    \item We developed a novel post-training graph convolution framework for extremely efficient training and devised a none-parametric GCN with self-loop, alleviating the embedding discrepancy issue.
    \item Originally, we proposed a continuous graph ordinary-differential-equation (LightGODE), which allows dynamic modeling of node representations and achieving optimal trade-offs of high-order information and embedding discrepancy.
    \item We conduct extensive experiments on three real-world datasets to test the effectiveness of LightGODE, demonstrating the highest recommendation performance with the lowest training time.
\end{itemize}

\section{Investigation of the Graph Convolution for Recommendation}
\label{sec: graph conv}

In this section, we initially investigate the necessity of graph convolution for recommendation and examine the key reasons behind the unexpectedly superior performance of the MF model enhanced by post-training graph convolution. Subsequently, we pinpoint the trade-offs in designing post-training graph convolution by identifying the embedding discrepancy issues when constructing deeper graph convolution layers.

\begin{figure}[htpb]
    \centering
    \includegraphics[width=0.88\linewidth]{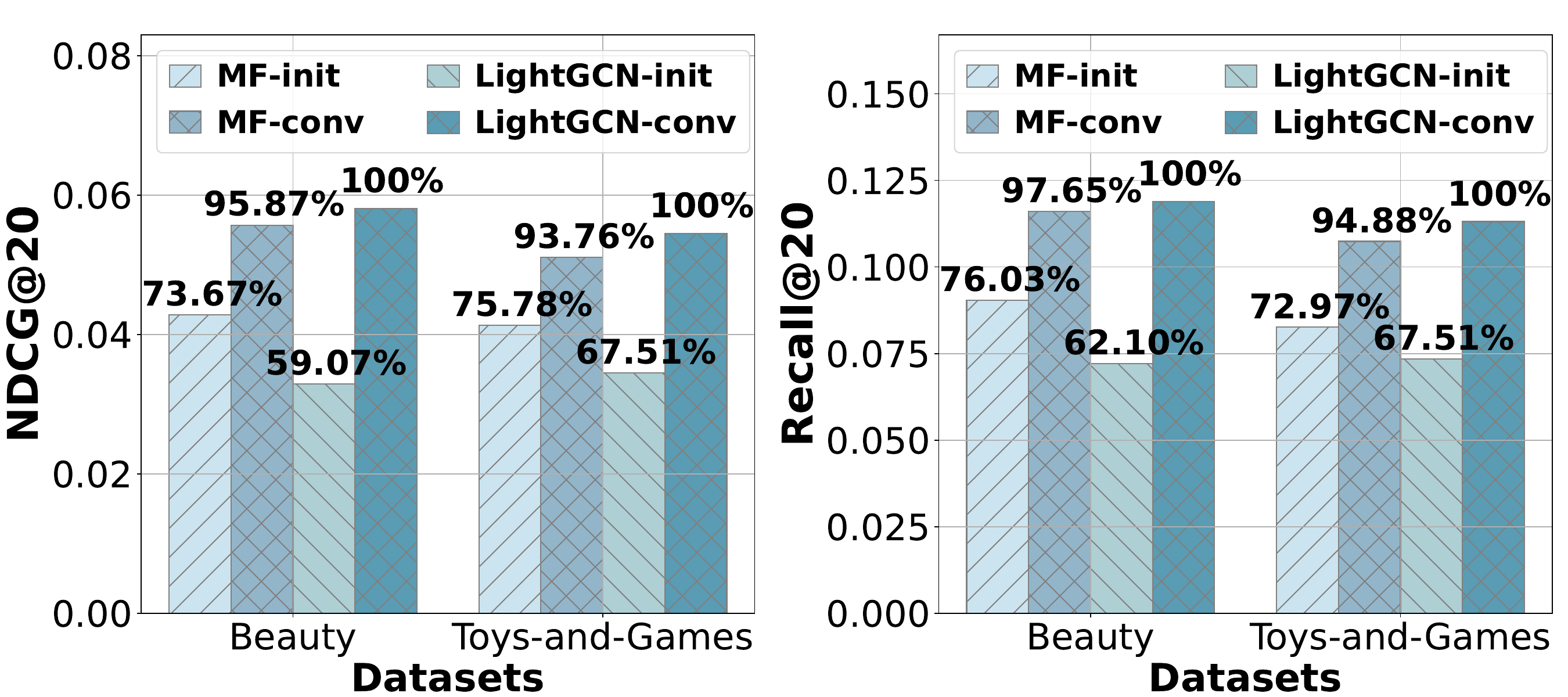}
    \caption{Preliminary study on the role of graph convolution for recommendation in training and testing stages. The MF model with graph convolution after training (MF-conv) achieves competitive results with the LightGCN-conv.}
    \label{fig: prelim}
\end{figure}

\vspace{-10pt}
\subsection{The Role and Necessity of Graph Convolution during Training}\label{sec: necessity}
To investigate the necessity of graph convolution for graph-based RecSys, we conduct preliminary experiments on the Amazon-Beauty (denoted as Beauty) and the Amazon-Toys-and-Games (denoted as Toys-and-Games) datasets to understand the impact of graph convolution in the training/testing stages of recommendation. Specifically, we design four variations of the model with the same amount of embedding parameters, including MF-init, which involves training with traditional matrix factorization and testing with its factorized ID embeddings; MF-conv, which integrates the 2-hops LightGCN convolution after MF training; LightGCN-init, which tests only with initial embeddings from a LightGCN model; and LightGCN-conv, which fully implements a 2-layer LightGCN model architecture.

As illustrated in Figure~\ref{fig: prelim}, we establish the LightGCN model (LightGCN-conv) as the benchmark by setting its performance as 100\%. To our surprise, MF-conv consistently outperforms both MF-init and Light-conv across the two datasets, achieving an impressive average of over 95\% of the performance metrics compared to LightGCN. This clearly highlights the substantial benefits of integrating post-training graph convolution with MF initialization, which significantly reduces computational costs by circumventing the intricate graph convolution process. Furthermore, these results underscore that the improved performance of graph-based RecSys primarily arises from the graph convolution after training, which prompts a thorough reconsideration of the necessity of the graph convolution during the training phase. Meanwhile, we propose a new point of view to understand the underlying reasons behind the unexpected exceptional performance of the MF-conv model even trained without graph convolution.

\subsection{The Alignment Force: A DFS Perspective}

Recommendation losses typically aim to identify the potential positive interactions via applying a supervised alignment force to positive user-item pairs during training. In this context, we empirically evaluated the alignment property \cite{wang2022towards, wang2020understanding} (the average distance between normalized positive embeddings) of four model versions in Section~\ref{sec: necessity} across the Beauty and Toys-and-Games datasets.

\begin{table} [ht]
\begin{center}
\caption{The alignment property of positive pairs in training.}
\label{tab: alignment}
\begin{tabular}{ccccc}
\toprule
\multirow{2}{*}{{Training}} & \multicolumn{2}{c}{{Beauty}} & \multicolumn{2}{c}{{Toys-and-Games}} \\
\cmidrule(r){2-3} \cmidrule(r){4-5}
& {Initial} & {Conv.} & {Initial} & {Conv.} \\
\midrule
MF   &0.7952&	0.6631&	0.8100&	0.7033\\
LightGCN   &0.8270&	0.6594&	0.7761&	0.6503\\
\bottomrule
\end{tabular}
\end{center}
\end{table}

In Table~\ref{tab: alignment}, the initial ID embeddings for both MF and LightGCN exhibit approximate alignment values in Beauty and Toys-and-Games, suggesting comparable training effects with and without lightweight graph convolution.
Similarly, the embeddings of post-training convolution show closely matched values, complying with the experimental findings of their comparable performance levels in Figure~\ref{fig: prelim}. These observations prompt us to explore whether the alignment forces exerted on user-item pairs, with and without the light graph convolution, are effectively equivalent.

Analytically, when a graph model is optimized by the same objective as the MF model, the alignment force applied on surrounding neighbors of positive pairs with graph convolution is the degree-weighted version of that alignment directly forced on two clusters of nodes. The assumption and proof are listed in the Appendix~\ref{append: align}.

\begin{figure}[htpb]
    \centering
    \includegraphics[width=0.9\linewidth]{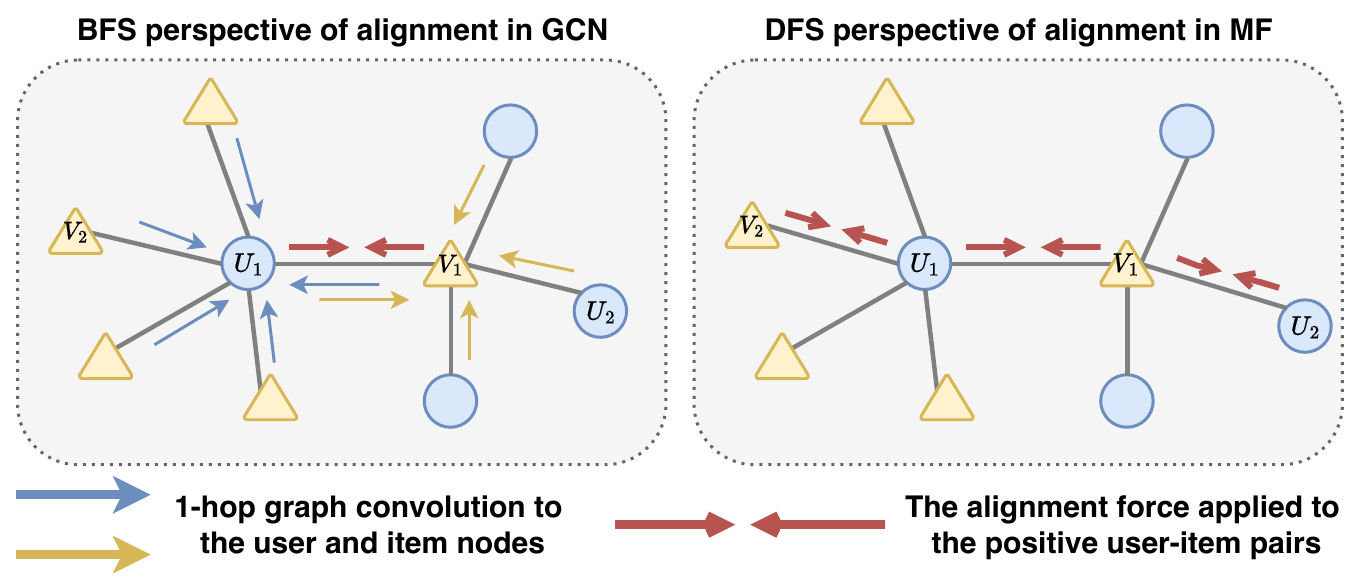}
    \caption{A comparison of alignment force in GCN-based and MF-based models from BFS and DFS, respectively. 
    }
    \label{fig: DFS}
\end{figure}

Intuitively, in an illustrative scenario where a single-layer graph convolution network is employed for gathering information, as depicted in the left section of Figure~\ref{fig: DFS}, one observes the subgraph comprising positive pairs $U_1$ and $V_1$ alongside their neighboring nodes. When the alignment force acts upon $U_1$ and $V_1$, the representations of their respective adjacent nodes, such as $V_2$ and $U_2$, also move closer together. Conversely, in the right portion of Figure~\ref{fig: DFS}, an alternative approach is showcased where, instead of employing BFS for neighborhood aggregation, direct connections between $U_1$ and nodes $V_2$ and $U_2$ are established via DFS paths among various user-item pairs, achieving a comparable outcome in terms of aligned representation learning. This conclusion further weakens the necessity of time-intensive graph convolution in training.

\subsection{Trade-off in Designing Graph Convolution}

\begin{figure}[htpb]
    \centering
    \includegraphics[width=0.92\linewidth]{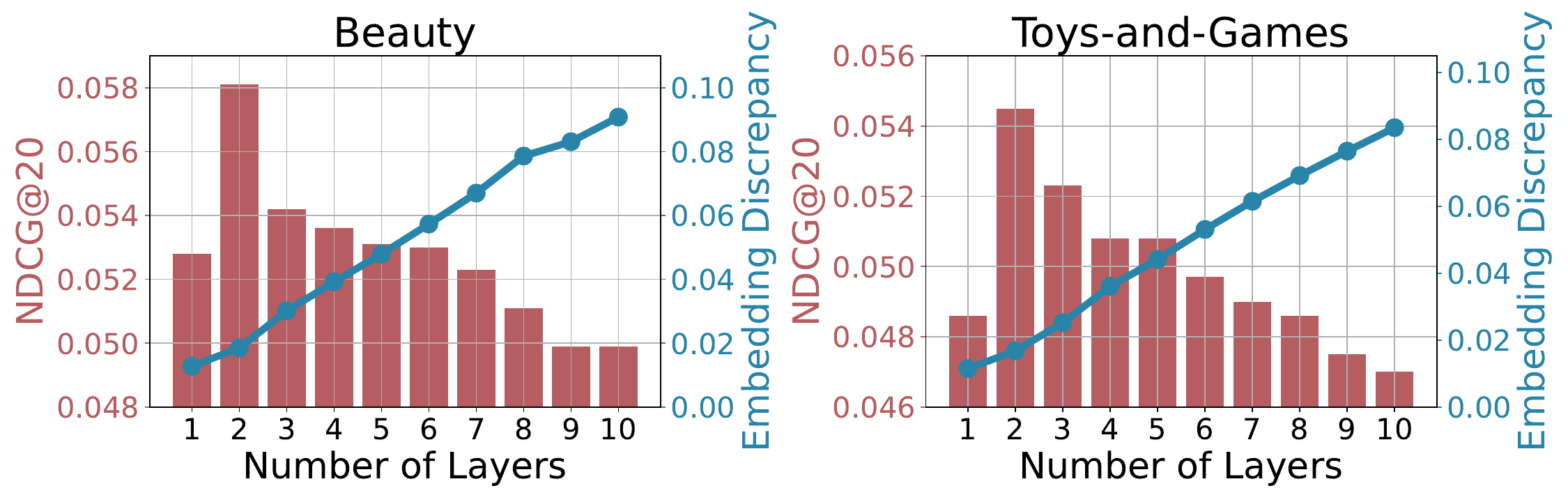}
    \caption{Study of the trade-off of embedding discrepancy and high-order information on Beauty and Toys-and-Games.}
    \label{fig: tradeoff}
\end{figure}

In the preceding sections, we highlighted the notable efficacy of the MF-conv model and analyzed the alignment properties to pinpoint key contributors to its approximate performance of LightGCN-conv. However, the MF-conv still slightly lags behind the LightGCN-conv, and it is not yet clear how to design a more effective non-parametric graph convolution in the post-training stage. 
The hint in the experiments (Figure~\ref{fig: prelim}) suggests that incorporating the multi-hop connectivity information during the testing phase proves beneficial. Then, it would be valuable to investigate whether higher-order connectivity continues to be advantageous after training with the MF model. 
In addition, if a model is optimally trained to fit the user-item interactions, one would expect the training embedding distribution to be ideal for testing. Consequently, any post-training operations should minimally impact the original embedding space. We are particularly interested in exploring how the model’s performance correlates with the differences between initial and convolution embeddings, termed as \textbf{Embedding Discrepancy}. We utilize the average Euclidean distance, widely acknowledged for numerical shifts \cite{goldenberg2019survey}, to quantify distribution shift across all users/items in the embedding space during graph convolution. 

In Figure~\ref{fig: tradeoff}, we empirically increase the layer number of post-training graph convolution, and the performance peaks at a two-hop convolution. 
Surprisingly, incorporating more complex, higher-order information mostly leads to a performance decline. 
Additionally, the discrepancy between initial and convolution embeddings enlarges with more layers, indicating that the existing graph convolution strategy can disrupt the foundational training, potentially causing over-smoothing \cite{rusch2023survey} as layers increase (Figure~\ref{fig: tradeoff}). 
This suggests that while additional convolution layers introduce more high-order information, they also risk perturbing well-trained embeddings. This could explain the counterexample as increasing the number of convolution layers initially promotes the performance and then continually brings negative effects - the current strategy finds a balance of configuring two layers.

To enhance performance, it is crucial to incorporate higher-order information while maintaining an embedding distribution close to that of the original MF model by adding more layers. This necessitates a more nuanced graph convolution approach that delicately constructs layers to maintain a trade-off between high-order structure information and the embedding discrepancy issue. In a comprehensive view of efficiency and effectiveness, we design a more fine-grained approach to balance the convolution depth and embedding discrepancy, as introduced in the following section.

\begin{figure*}[htpb]
    \centering
    \includegraphics[width=1.0\linewidth]{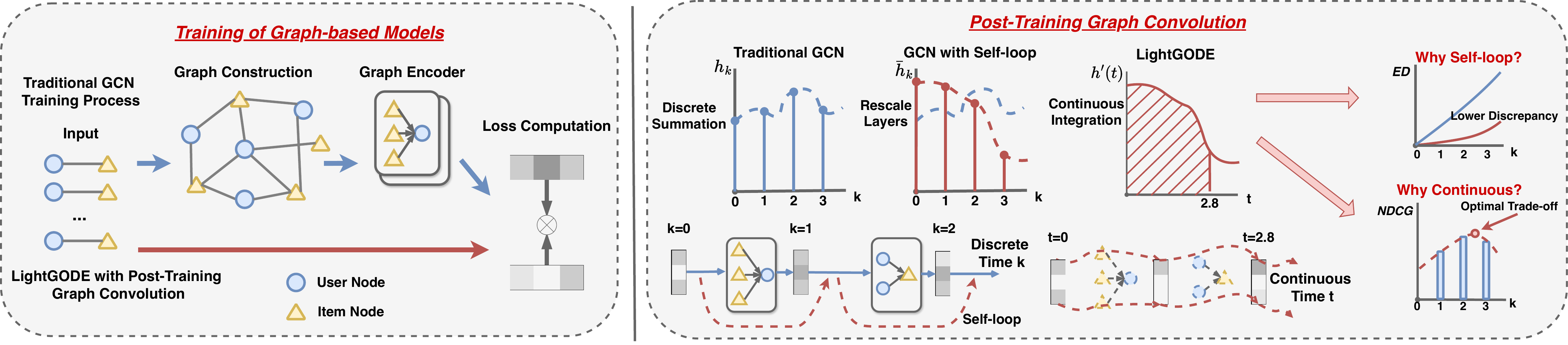}
    \caption{The training pipeline of traditional GCN-based recommendation and our proposed LightGODE with post-training graph convolution (PTGC) framework, where we skip the time-consuming convolution-related operations to speed up the training. In the PTGC stage, the self-loop prioritizes the shallow layers by weighing more on preceding layer representations, thus mitigating the distribution discrepancy problem. Based on the design of discrete non-parametric GCN, we derive LightGODE, a continuous ODE function that implements fine-grained graph convolution to achieve the optimal trade-off in the GCN design.}
    \label{fig: main}
\end{figure*}

\section{Light Post-Training Graph-ODE for Efficient Recommendation}
In this section, we propose the post-training graph convolution framework, including the pre-training user/item embeddings for extremely efficient graph recommendation. To balance the integration of high-order information and the risk of embedding discrepancy, we devise a non-parametric graph convolution with self-loop. Based on the formulation, we propose that LightGODE - a continuous post-training graph convolution based on ordinary-differential-equations aiming to achieve the optimal trade-off. Finally, a detailed time complexity analysis and comparison with other strong GCN baselines are demonstrated.

\subsection{Pre-training User/Item Embedding} \label{sec: train}
Here, we outline our overall training pipeline toward the extremely efficient graph-based recommendation. Since the graph convolution proved unnecessary in the training stage as Section~\ref{sec: graph conv}, we abandoned the graph convolution-related operations and focused solely on training the randomly initialized ID embeddings,  as shown in the training section of Figure~\ref{fig: main}. Regarding the loss computation, we directly optimize the alignment and uniformity properties as in \cite{wang2022towards} to reach optimal status for MF embedding training as an ideal foundation for the subsequent graph convolution phase. Specifically, the alignment loss minimizes the distance between the normalized embeddings of the positive pairs $(\mathbf{u}_i, \mathbf{v}_j)$ within batch $\mathcal{B}$:
\begin{equation}
\mathcal{L}_{align}=\frac{1}{|\mathcal{B}|} \sum_{(\mathbf{u}_i,\mathbf{v}_j) \in \mathcal{B}} \left\| \mathbf{u}_i-\mathbf{v}_j\right\|^2.
\end{equation}
The uniformity loss $\mathcal{L}_{uniform}={(\mathcal{L}^U_{uniform}+\mathcal{L}^V_{uniform})}/2$, and the user-side uniformity is given by: 
\begin{equation}
\mathcal{L}^U_{uniform}=\log \frac{1}{|\mathcal{B}_u|^2} \sum_{\mathbf{u}_i \in \mathcal{B}_u} \sum_{\mathbf{u}_{i^\prime} \in \mathcal{B}_u} e^{-2 \left\| \mathbf{u}_i-\mathbf{u}_{i^\prime} \right\|},
\end{equation}
where $\mathcal{B}_{u}$ is the user batch and $\mathbf{u}_{i^\prime}$ is rest of users in batch. The item side uniformity $\mathcal{L}^U_{uniform}$ follows the same format and final loss becomes $\mathcal{L} = \mathcal{L}_{align} + \gamma \mathcal{L}_{uniform}$ adjusted by weight $\gamma$.

\subsection{Discrete GCN with Self-Loop}
Empirical evidence in \cite{wang2020disentangled, he2020lightgcn} and Section~\ref{sec: graph conv} suggests that optimal performance is typically achieved when the graph model is configured with two or three layers. However, abruptly discontinuing the convolution process at higher-order layers is inappropriate since neither the preceding shallow layers are distinctively treated nor the subsequent high-order layers are noticed. Such an approach lacks a seamless transition from lower to higher-order graph convolutions, potentially overlooking nuanced differences in structural information embedded in shallow and deep graph relationships. This calls for reconsidering the graph convolution process across different layer depths to better capture the complexity and dynamics of the graph data in recommendation contexts.

One straightforward solution is to integrate the self-loop (SL) into the graph convolution process. This simple operation highlights the importance of the node representations of preceding layers in each message-passing process, contributing to a gradual transition to higher-order connectivity. 
Suppose we observe a pair of interacted users and items with corresponding initial input ID embeddings $\mathbf{u}^0_i$ and $\mathbf{v}^0_j$, and we design the parameter-free graph convolution based on the smoothed neighborhood aggregation process as in \cite{he2020lightgcn}. The graph convolution with SL is finalized as:
\begin{equation}\label{eq: discrete}
\begin{aligned}
\mathbf{u}^{k}_i = \mathbf{u}^{k-1}_i + \sum_{j\in N_i} \frac{1}{\sqrt{|N_i|} \sqrt{|N_j|}}\mathbf{v}^{{k-1}}_j ,\\
\mathbf{v}^{k}_j = \mathbf{v}^{k-1}_j + \sum_{i\in N_j} \frac{1}{\sqrt{|N_j|} \sqrt{|N_i|}}\mathbf{u}^{k-1}_i,\\
\end{aligned}
\end{equation}
where $\mathbf{u}^{k-1}_i$ and $\mathbf{v}^{k-1}_j$ are embeddings of user $\mathbf{u}_i$ and item $\mathbf{v}_j$ at layer of $k-1$, respectively. The normalization employs the average degree $\frac{1}{\sqrt{|N_i|} \sqrt{|N_j|}}$ to temper the magnitude of popular nodes after graph convolution.
Afterward, the collaborative filtering final embedding is obtained by synthesizing the layer-wise representations:
\begin{equation}
\begin{aligned}
\mathbf{u}_i^{(K)} = \sum^{K}_{k=0}  \mathbf{u}^k_i;\quad
\mathbf{v}_j^{(K)} = \sum^{K}_{k=0}  \mathbf{v}^k_j.
\end{aligned}
\end{equation}

\subsection{Continuous Graph-ODE}
Motivated by \cite{chamberlain2021grand, thorpe2022grand++, xhonneux2020continuous} deriving continuous differential equations from diffusion process to model the dynamics in the graph, we aim to design a continuous version of our discrete non-parametric graph convolution with self-loops.

Formally, given $\mathbf{h}_{0}$ as the initial embedding of the users and items, we can rewrite the layer-wise information update Equation~\ref{eq: discrete} in terms of the matrix operations:
\begin{equation}\label{eq: propogation}
\mathbf{{h}}_{k}=\mathbf{{A}}{\mathbf{h}_{k-1}},
\end{equation}
where $\mathbf{h}_{k}$ is the node embeddings of $k$-th layer, aggregating their neighborhood information and fusing with its own representation of the previous layer via self-loop.
The matrix $\mathbf{A}=\mathbf{\bar{A}} + \mathbf{I}$ and $\mathbf{\Bar{A}} $ is the normalized adjacency matrix.

Consequently, the end result for a $K$-layer discrete graph convolution network $\mathbf{h}{(K)}$ can be represented as:
\begin{equation}\label{eq: feature}
\mathbf{h}{(K)}= \sum_{k=0}^{K}{\mathbf{h}_{k}}
= \sum_{k=0}^{K}{{\mathbf{A}}^k {\mathbf{h}_{0}}}.
\end{equation}

This sum from Equation~\ref{eq: feature} can be seen as a Riemann sum extending from layer $0$ to layer $K \rightarrow \infty$, transitioning to a continuous ODE function (proof provided in the Appendix~\ref{append: ode}):
\begin{equation}
\frac{\mathrm{d} \mathbf{h}(t)}{\mathrm{d} t}=\ln \mathbf{A} \mathbf{h}(t) + (\mathbf{A} - \ln \mathbf{A}) \mathbf{h}_0,
\end{equation}
which simplifies under a first-order Taylor expansion approximation where $\ln \mathbf{A} = \mathbf{A} - \mathbf{I} = \mathbf{\bar{A}}$, leading to: 
\begin{equation}
\frac{\mathrm{d} \mathbf{h}(t)}{\mathrm{d} t}=\mathbf{\bar{A}} \mathbf{h}(t) 
+\mathbf{h}_0.
\end{equation}

The general form of this continuous graph convolution network is obtained by integration from the initial condition as:  
\begin{equation}\label{eq: cgnn}
 \mathbf{h}(t)= h_0 + \int_0^{t}[\mathbf{\bar{A}} \mathbf{h}(s)   + \mathbf{h}_0] \mathrm{d}(s).
\end{equation}
Note that the final integration form could be solved analytically using the integration factor. However, considering that computing the exponential of the matrix in the analytical solution is time-consuming, we resort to the simple and fast Euler solver \cite{chen2018neuralode} to approximate the ODE solution.

\subsection{Time Complexity Analysis}
In this subsection, we analyze the computation complexity of LightGODE and compare it with two prominent GCN benchmark methods, LightGCN \cite{he2020lightgcn} and GraphAU \cite{yang2023graph}. We first define the number of edges in the user-item bipartite graph as $|\mathcal{E}|$. Then, let $K$ represent the number of graph convolution layers and $d$ the size of embeddings.
On this basis, we can derive the following facts: 
\begin{itemize}[leftmargin=*]
    \item In the graph construction process, both LightGCN and GraphAU require normalization of the adjacency matrix. This step involves computing $2|\mathcal{E}|$ non-zero elements of the original adjacency matrix. On the contrary, LightGODE alleviates the need for graph construction and adjacency matrix normalization in training.
    \item In the graph convolution stage, LightGCN and GraphAU both perform linear message-passing through the graph's edges in each layer, which incurs a computational cost of $2|\mathcal{E}|Kd$. Whereas LightGODE does not involve graph convolution in training, significantly facilitating large-scale graph recommendations.
    \item Regarding the loss computation, LightGCN adopts the BPR loss for optimization, leading to a computational demand of $O(2Bd)$. GraphAU, on the other hand, uses alignment and uniformity loss calculations between users and items in the batch, resulting in a time complexity in the batch of $O(2KBd + 2B^2d)$. LightGODE, focusing only on the alignment loss at the initial embedding, has a time complexity per batch of $O(Bd + 2B^2d)$. It should be noted that all the experiments are implemented on GPU-based parallel computation, which minimizes the relative importance of batch size $B$ in model comparisons. Furthermore, the BPR loss's reliance on negative sampling for each user-item pair in every batch through all epochs makes LightGCN less efficient than LightGODE in handling large-scale graphs.
\end{itemize}

\begin{table}[ht]
\small
\centering
\caption{Time complexity comparison of LightGCN, GraphAU, and LightGODE during training.}

\begin{tabular}{m{1.65cm} |m{1.5cm}|m{2.1cm}|m{1.9cm}}
\toprule
 Stages & LightGCN & GraphAU & LightGODE\\
\midrule
\makecell[l]{Adjency \\Matrix} & \( O(2|\mathcal{E}|) \) & \( O(2|\mathcal{E}|) \) & - \\
\midrule
\makecell[l]{Graph \\ Convolution} & \( O(2|\mathcal{E}|Kd) \) & \( O(2|\mathcal{E}|Kd) \)  & - \\
\midrule
\makecell[l]{Loss\\ Computation}   & \( O(2Bd) \) & \( O(2KBd+2B^2d) \) & \( O(Bd+2B^2d) \)\\

\bottomrule
\end{tabular}
\label{tab:complexity}
\end{table}

\vspace{-10pt}
\section{Experiments}
\subsection{Datasets}

\begin{table*}[htpb]
\small
\begin{center}
\caption{Performance comparison on three benchmark datasets in terms of NDCG and Recall.}
  \label{tab:main}
  \begin{tabular}{ccccccccccccc}
    \toprule
    \multirow{2}{*}{\textbf{Method}} & \multicolumn{4}{c}{\textbf{Gowalla}} & \multicolumn{4}{c}{\textbf{Beauty}}  & \multicolumn{4}{c}{\textbf{Toys-and-Games}}\\
    
    \cmidrule(r){2-5} \cmidrule(r){6-9} \cmidrule(r){10-13} 
    & \textbf{N@20} & \textbf{R@20} &\textbf{ N@50} & \textbf{R@50} 
    & \textbf{N@20} & \textbf{R@20} &\textbf{ N@50} & \textbf{R@50} 
    & \textbf{N@20} & \textbf{R@20} &\textbf{ N@50} & \textbf{R@50}  \\
    
    \midrule
    BiasMF & 0.0406 & 0.0700 & 0.0507 & 0.1122 
    &0.0428 & 0.0904 & 0.053 & 0.1404 
    & 0.0413 & 0.0826 & 0.0503 & 0.1271\\ 
    NeuMF & 0.0487 & 0.0952 & 0.0637 & 0.1597 
    & 0.0343 & 0.0746 &  0.043 & 0.1173 
    & 0.0301 & 0.0632 & 0.0375 & 0.0994\\
    
    NGCF & 0.0501 & 0.0923 & 0.0644 & 0.1535 
    & 0.0438 & 0.0943 &  0.0559 & 0.1537 
    & 0.0379 & 0.0827 & 0.0486 & 0.1356\\
    DGCF & 0.0553  & 0.0967 & 0.0692 & 0.1556 
    & 0.0516 & 0.1081 & 0.0624 & 0.1610 &
    0.0485 & 0.1007 & 0.0589 & 0.1515\\

    SimpleX & 0.0451 & 0.0876 & 0.0611 & 0.1555 
    & 0.0502 & 0.1104 & 0.0623 & 0.1697
    & 0.0521 & 0.1092 & 0.0632 & 0.1640\\
    LightGCN & 0.0683 & 0.1224 & 0.0860 & 0.1974 
    & 0.0581 & 0.1189 & 0.0709 & 0.1816
    & 0.0555 & 0.1131 & 0.0669 & 0.1696\\
    ODE-CF & 0.0680	& 0.1220 & 0.0854 &	0.1960 
    &0.0537 & 0.1158 &0.0661 &	0.1760
    & 0.0516 & 0.1075 & 0.0633 & 0.1656\\

    \midrule
    DirectAU & 0.0768	& 0.1437 & 0.0978 &	0.2319
    & 0.0555 & 0.1149 & 0.0673 & 0.1725
    & 0.0571 & 0.1184 & 0.0677 & 0.1714\\
    GraphAU & \underline{0.0811} & \underline{0.1461} & \underline{0.1017} & \underline{0.2346} 
    & \underline{0.0662} & \underline{0.1398} & \underline{0.0782} & \underline{0.2116}
    & \underline{0.0622} & \underline{0.1324} & \underline{0.0725} & \underline{0.1952 }\\
    \midrule
    
    LightGODE & \textbf{0.0929} & \textbf{0.1678} &  \textbf{0.1150} & \textbf{0.2628} 
    & \textbf{0.0714} & \textbf{0.1452} & \textbf{0.0852} & \textbf{0.2130} 
    & \textbf{0.0673} & \textbf{0.1371} & \textbf{0.0794} & \textbf{0.1983} \\
  \bottomrule
\end{tabular}
\end{center}
\end{table*}

We experiment on three public real-world datasets: Gowalla, Amazon-Beauty (Beauty), and Amazon-Toys-and-Games (Toys-and-Games), varying in scales and domains.  
The Gowalla~\footnote{\url{https://snap.stanford.edu/data/loc-gowalla.html}} is a location-based social networking dataset obtained from users' checking-in.
Beauty and Toys-and-Games are crawled from real-world data in Amazon~\footnote{\url{https://jmcauley.ucsd.edu/data/amazon/links.html}} according to the product category. 
We follow the 5-core setting in \cite{wang2022towards, zhang2024mixed} by removing the users/items with node degrees less than five to ensure the data quality for testing. 
We split all datasets into training (80\%), validation (10\%), and testing (10\%), and the statistical information of the three datasets after filtering is summarized in Table ~\ref{tab: data}. More details about implementations, evaluations, and baseline can be found in Appendix \ref{append: experiment}.

\begin{table}[ht]
\small
\begin{center}
\caption{The statistics of the datasets.}
  \label{tab: data}
  \begin{tabular}{ccccc}
    \toprule
    Dataset & \# Users & \# Items & \# Interactions & Sparsity\\
    \midrule
    Gowalla & 64,116 & 164,533 & 2,018,421 & 99.9809\%\\
    Beauty & 22,364 & 12,102 & 198,502 &	99.9267\%\\
    Toys-and-Games &  19,413 & 11,925 & 167,597 & 99.9276\%\\
  \bottomrule
\end{tabular}
\end{center}
\end{table}

\subsection{Overall Performance Comparison}

In this comprehensive experiment, we compare the performance of several state-of-the-art recommendation algorithms on three diverse datasets: Gowalla, Beauty, and Toys-and-Games, using NDCG20, NDCG@50, Recall@20, and Recall@50. Here, we highlight the main observations as follows:

\begin{itemize}[leftmargin=*]
\item Noticeably, LightGODE achieves the highest scores in NDCG and Recall across all datasets, demonstrating its effectiveness in different recommendation tasks. It should highlighted that in the large-scale dataset Gowalla (with 64,116 users and 164,533 items), LightGODE surpasses all the other baseline methods by large margins with more than 10\% improvement over the strongest baseline, emphasizing its potential to be deployed in the large-scale graphs in real-world applications.
\item Among all, DirectAU and GraphAU emerge as the most competitive baselines in all three datasets, demonstrating the effectiveness of the alignment and uniformity \cite{wang2020understanding} in optimization.
\item Most of the graph-based recommender systems consistently outperform the traditional MF models. This suggests the importance of graph convolution for capturing the multi-hop information. Though leveraging contrastive learning loss, SimpleX performs poorly in the context of sparse dataset Gowalla, whereas LightGCN and ODE-CF are more robust across all datasets.
\end{itemize}

\subsection{Ablation Study}
Ablation studies on LightGODE are conducted to validate the rationality and effectiveness of our design choices. From Table (\ref{tab: ablation}), it is evident that the full version of the LightGODE model achieves the best scores across all metrics and datasets, showcasing the efficacy of the continuous ODE function. Furthermore, using only our parameter-free graph convolution with self-loop (w/o ODE) still results in higher NDCG and Recall and lower embedding discrepancy compared to traditional lightweight graph convolution (w/o SL), indicating more consistent embeddings. The model without post-training graph convolution (w/o Conv) exhibits the lowest performance. Therefore, each component within our LightGODE contributes significantly to the final recommendation performance.

\begin{table} [ht]
\small
\begin{center}
\caption{Abalation study on different components. The embedding discrepancy (ED) is the Euclidean distance between initial and convolution embeddings; the lower the better.}
\label{tab: ablation}
\begin{tabular}{m{1.3cm}|m{1.1cm}|m{1.0cm}m{1.0cm}m{1.0cm}m{1.0cm}}
\toprule
Dataset & Metrics & \textbf{\makecell[l]{Light-\\GODE}} & \textbf{\makecell[l]{w/o \\ODE}} &\textbf{ \makecell[l]{w/o \\SL}} & \textbf{\makecell[l]{w/o \\Conv}} \\
\midrule
\multirow{2}{1cm}{{{{Gowalla}}}} 
& {NDCG} & \textbf{0.0929} &	0.0833 &	0.0801 &	0.0768 \\
& {Recall} & \textbf{0.1678} &	0.1537 &	0.1481 &	0.1437 \\
& {ED} & \textbf{0.0066} &	0.0158 &	0.0282 &	- \\
\midrule
\multirow{2}{1cm}{{{{Beauty}}}} 
& {NDCG} & \textbf{0.0714}&	0.0700 &	0.0686 &	0.0555 \\
& {Recall} & \textbf{0.1452}&	0.1450 &	 0.1428 &	0.1149 \\
& {ED} & \textbf{0.0022} &	0.0049 &	0.0082 &	- \\
\midrule
\multirow{2}{1.5cm}{{{{Toys-and-Games}}}} 
& {NDCG} & \textbf{0.0673}&	0.0644&	 0.0641&	0.0571 \\
& {Recall} & \textbf{0.1371}&	 0.1343& 0.1337&	0.1184 \\
& {ED} & \textbf{0.0011} &	0.0045 &	0.0067 &	- \\
\bottomrule
\end{tabular}
\end{center}
\end{table}

\subsection{Efficiency Analysis}
\subsubsection{\textbf{Trade-off between the Performance and the efficiency.}}
\begin{figure}[htpb]
    \centering
    \includegraphics[width=0.8\linewidth]{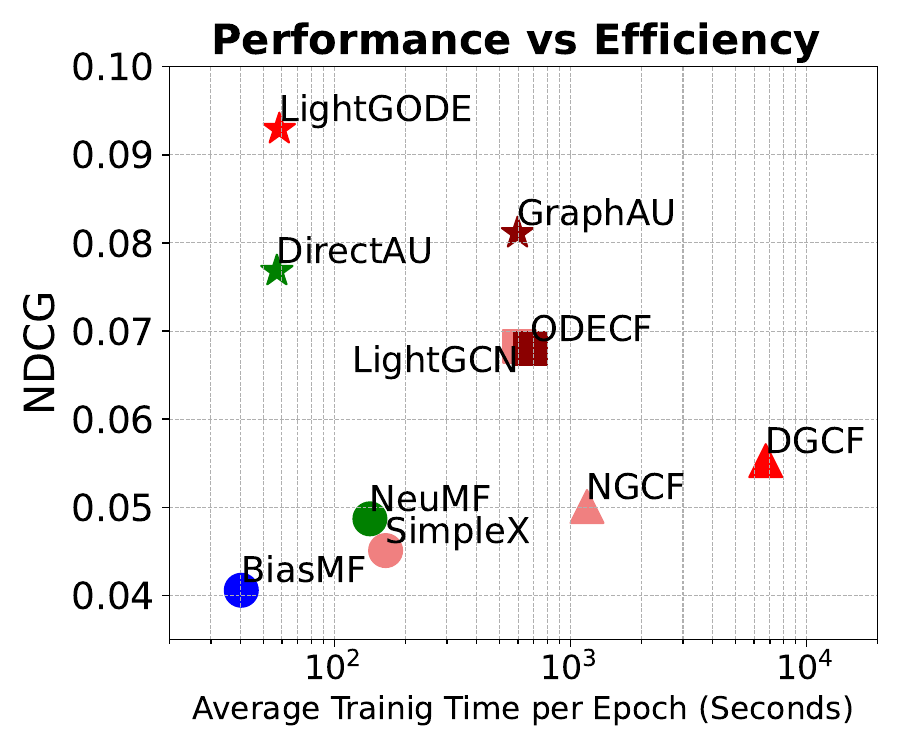}
    \caption{Trade-off between the performance and the efficiency on the Gowalla dataset. The left upper direction indicates stronger performance and more efficient training.}
    \label{fig: efficiency}
\end{figure}

Figure \ref{fig: efficiency} illustrates the overall comparison of performance and efficiency on the large Gowalla dataset. LightGODE markedly outperforms all benchmarks while maintaining high efficiency, underscoring its potential for effective, large-scale recommendation systems.
Early works that leverage GCN encoders, such as NGCF and DGCF, fall behind in average training times per epoch and NDCG. More advanced GCN-based approaches, including LightGCN, ODECF, and GraphAU, show substantial improvements in NDCG scores yet are still much slower than simpler MF models in speed. Conversely, BiasMF, NeuMF, and SimpleX directly utilize user-item interactions, achieving notably low training times but exhibiting poor ranking scores. Only DirectAU manages a balanced trade-off but still lags behind LihgtGODE regarding NDCG.

\subsubsection{\textbf{Training Time Comparison.}}
To delve deeper into the efficiency and scalability analysis in terms of the training, we provide a comprehensive training time comparison featuring the average time per epoch, the number of required epochs, and the total training cost shown in Table \ref{tab: time}. On the large dataset Gowalla, NGCF and DGCF consume longer times per epoch for training, taking tens of hours in total to reach the optimal status. LightGCN and ODECF demonstrate shorter epoch duration but demand a greater number of epochs to complete training. Although GraphAU exhibits the fastest training speed per epoch among the baseline methods in the Beauty and Toys-and-Games dataset, it is almost as slow as LightGCN and ODECF, especially on the large Gowalla dataset. LightGODE significantly reduces the overall training time to less than one hour on Gowalla. These observations highlight the efficiency and scalability of LightGODE towards industrial RecSys.

\begin{table}[htpb] 
\small
\begin{center}
\caption{Training time comparison of GCN-based models on Gowalla, Beauty, and Toys-and-Games datasets. It includes the average training time per epoch, the number of epochs, and the total training time. For abbreviation, we denote seconds as s, minutes as m, and hours as h.}
\label{tab: time}
  \begin{tabular}{cc|ccc}
    \toprule

Dataset  & Method  & Time/Epoch & \# Epochs & Total Time \\
\midrule
\multirow{6}{*}{\textbf{Gowalla}} 
&NGCF    &1175.79s	&84&	27.44h \\
&DGCF         &6720.88s	&43&	80.28h \\
&LightGCN        &608.69s	&105&	17.75h \\
&ODECF     &679.25s	&79&	14.91h \\
&GraphAU     &597.11s	&91&	15.09h \\
&LightGODE     &58.46s	&61&	\textbf{0.99h} \\
\midrule
\multirow{6}{*}{\textbf{Beauty}} 
&NGCF    &10.79s	&48&	8.63m \\
&DGCF         &204.95s	&72&	245.94m \\
&LightGCN        &6.76s	&83&	9.35m \\
&ODECF     &8.34s	&108&	15.01m \\
&GraphAU     &8.72s	&41&	5.96m \\
&LightGODE     &3.16s	&68&	\textbf{3.58m} \\
\midrule
\multirow{6}{*}{\makecell[l]{\textbf{Toys-and}\\\textbf{-Games}}} 
&NGCF    &11.19s	&57&	10.63m \\
&DGCF         &116.69s	&61&	118.64m \\
&LightGCN        &5.01s	&119&	9.94m \\
&ODECF     &6.60s	&147&	16.17m \\
&GraphAU     &7.09s	&41&	4.84m \\
&LightGODE     &2.65s	&76&	\textbf{3.36m} \\
  \bottomrule
\end{tabular}
\end{center}
\end{table}

\subsubsection{\textbf{Performance Curve and Convergence Speed}}

\begin{figure}[htpb]
    \centering
    \includegraphics[width=0.9\linewidth]{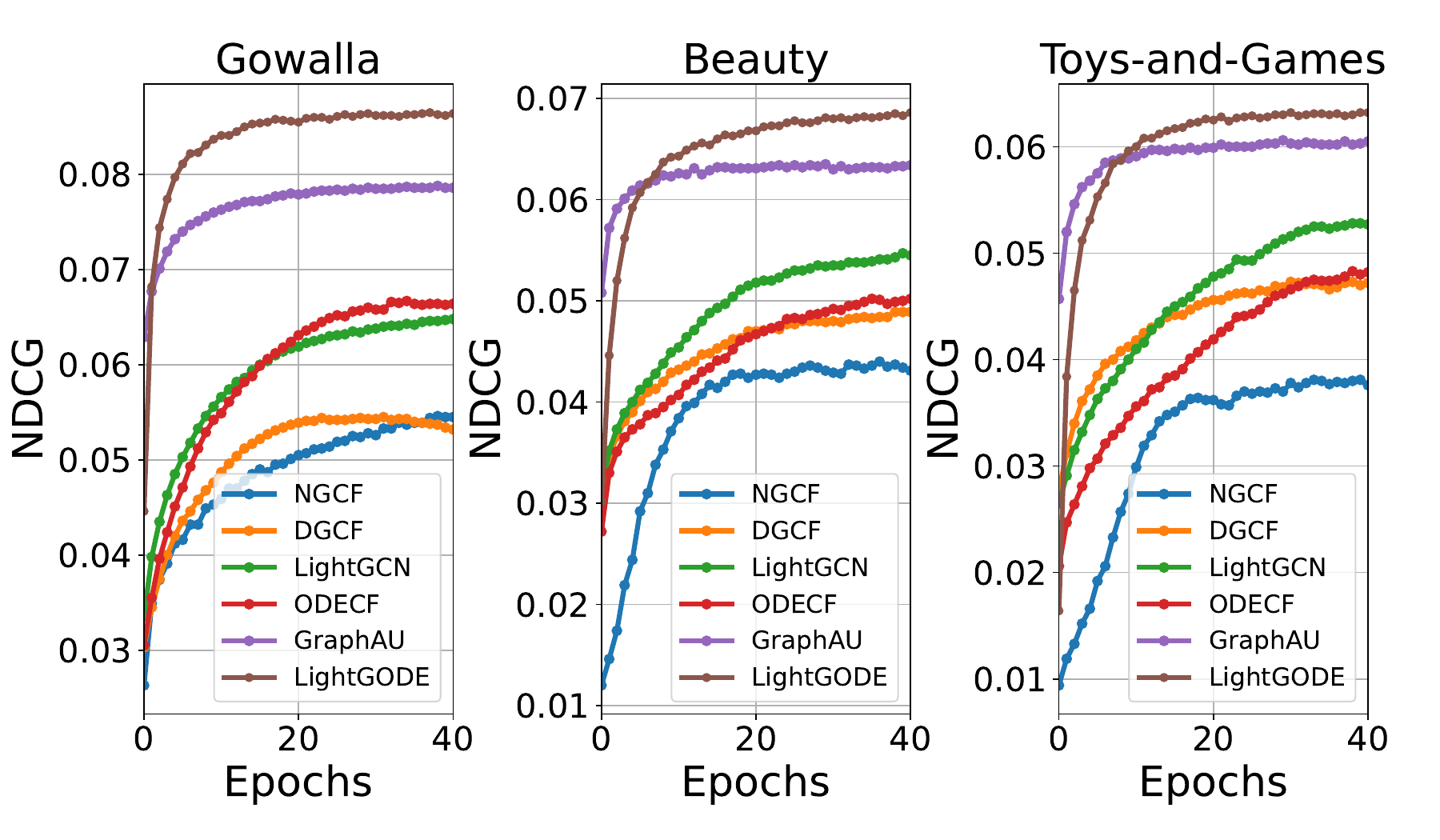}
    \caption{Performance curve in the first 40 epochs.}
    \label{fig: curve}
\end{figure}

In Figure \ref{fig: curve}, we present the training curves of performance against epochs on the tree datasets. Overall, NGCF and DGCF exhibit low-performance peaks, while LightGCN and ODECF converge slowly. By enforcing alignment and uniformity in the representation hyperspace, the performances of GraphAU and LightGODE ensure convergence at early training stages. Our method requires fewer epochs to converge and consistently results in high recommendation scores.

\subsection{Comparison with GODE}
\begin{figure}[htpb]
    \centering
    \includegraphics[width=0.8\linewidth]{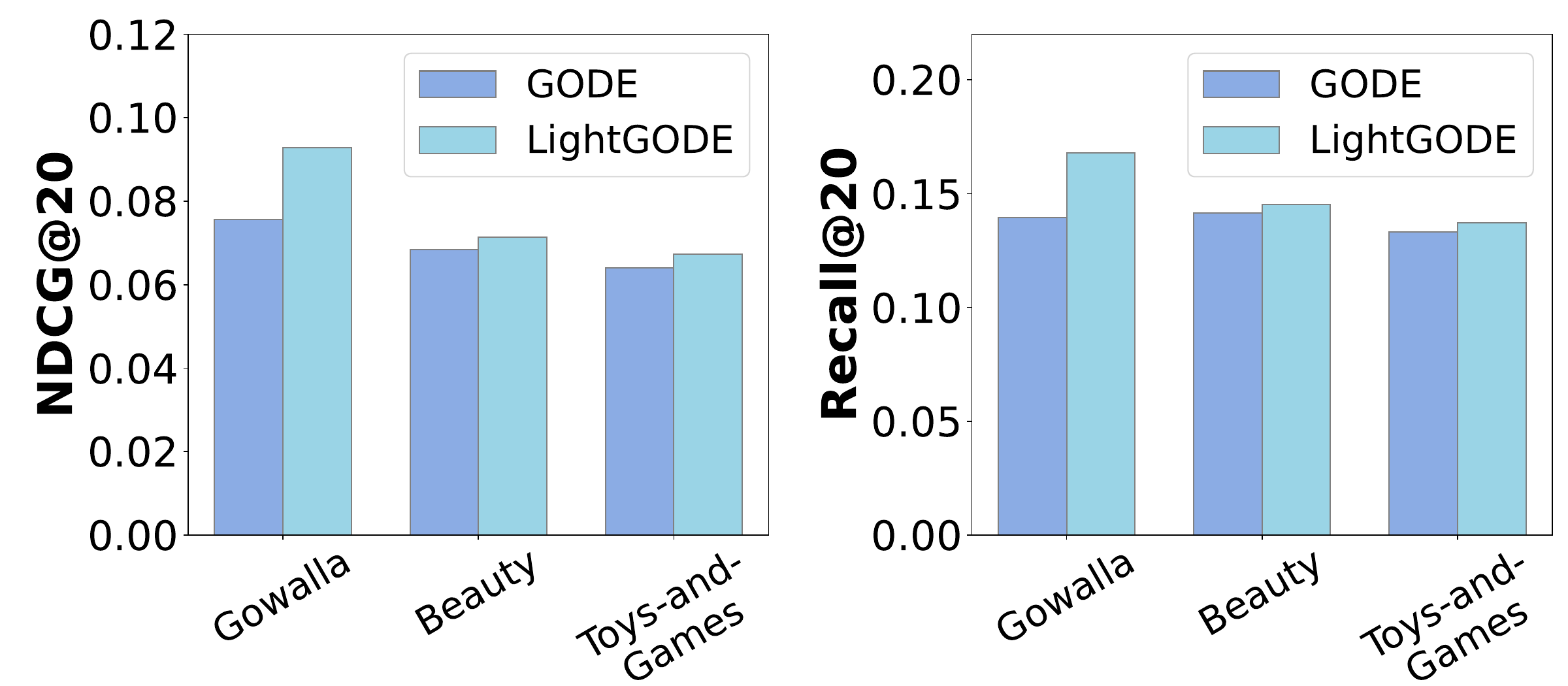}
    \vspace{-8pt} 
    \caption{Performance Comparison of LightGODE and GODE.}
    \label{fig: sparse_beauty}
\end{figure}
To evaluate the effectiveness of our continuous ODE function and self-loop operations tailored for post-training graph convolution, we compare LightGODE with post-training graph convolution and GODE with pre-training graph convolution. LightGODE consistently emerges as the superior performer across all metrics in all three datasets, especially in the large-scale dataset Gowalla. This demonstrates that our innovative design for post-training graph convolution not only allows our simple model to exceed the performance of GCN models trained with traditional graph convolution but also significantly speeds up the training strategy.

\subsection{Hyperparameter Analysis}
\subsubsection{\textbf{Impact of the Time $t$}}
\begin{figure}[htpb]
    \centering
    \includegraphics[width=1.0\linewidth]{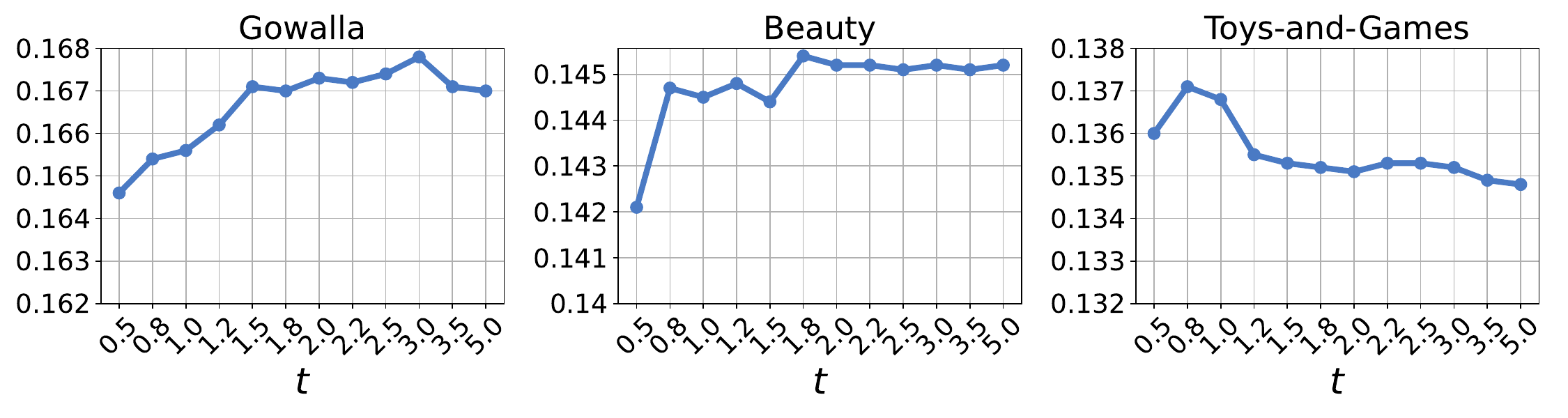}
    \vspace{-20pt} 
    \caption{Impact the time $t$ on Recall.}
    \label{fig: tem}
\end{figure}

We evaluate how the continuous time $t$ affects the performance of LightGODE. As observed in Figure~\ref{fig: tem}, it suggests that an appropriate time $t$ is generally associated with the scale of the datasets. In Gowalla, performance peaked at around 3 as a larger time $t$ enables the model to long-distance neighborhood aggregations. Whereas in Beauty and Toys-and-Games, $t$ is chosen at 1.8 and 0.8, indicating a smaller receptive field to achieve the optimal convolution depth.

\subsubsection{\textbf{Impact of the uniformity weight $\gamma$}}
Another hyperparameter is the weight of the uniformity loss $\gamma$. From the curves in Figure 9, a smaller uniformity weight (0.5) achieves the highest recall on the Beauty and Toys-and-Games datasets. In contrast, larger values of $\gamma$ are detrimental to the recommendation performance in Gowalla datasets. As for large-scale datasets, the user and item representations should be more evenly distributed so as to make the user/item embeddings more representative for distinction.

\begin{figure}[htpb]
    \centering
    \includegraphics[width=0.95\linewidth]{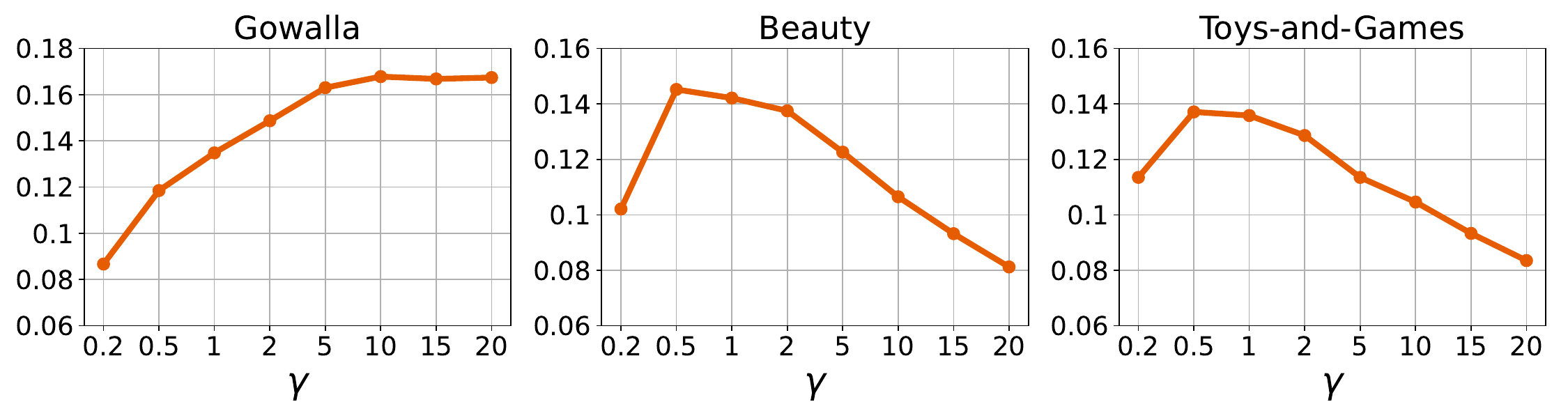}
    \vspace{-10pt} 
    \caption{Impact of the uniformity weight $\gamma$ on Recall.}
    \label{fig: n_mix}
\end{figure}

\section{Related Work}
\subsection{Graph Convolution Network for RecSys}
Collaborative filtering (CF) is widely used to provide recommendations based on user-item interactions. Recent developments in graph convolution networks (GCNs) have reformed CF from conventional matrix factorization CF to Graph-based CF, incorporating social networks \cite{fan2019graph, liu2022federated, yang2021consisrec}, knowledge graph \cite{wang2019kgat, cao2019unifying, liu2024knowledge}, and user-item interactions \cite{wang2019neural, ying2018graph, wang2020disentangled, he2020lightgcn, mao2021ultragcn, yang2022graph, zhang2023dual}. One of the early attempts is NGCF \cite{wang2019neural}, which incorporates the importance of high-order connectivity in the user-item bipartite graph. Another early work, PinSAGE \cite{ying2018graph}, utilizes random-walk to sample subgraphs and scales up RecSys industrial level. DGCF \cite{wang2020disentangled} disentangles latent intentions of users and diversifies item recommendations, thus yielding better performance and interpretability. LightGCN \cite{he2020lightgcn} is a lightweight framework that omits linear transformations and nonlinear activations in GCN layers and drastically improves the efficiency of graph recommendation. To further simplify the graph training process, UltraGCN \cite{mao2021ultragcn} adjusts the relative importance of nodes to aggregate the embeddings by weights and directly approximates the converged state of message passing, which accelerates LightGCN by more than multiple times. GraphAU \cite{yang2023graph} identifies the inefficiency of DirectAU \cite{wang2022towards} on graph-based recommendations and proposes high-order representation alignment for the linear scale of computation for additional layers. ODECF \cite{xu2023graph} condenses multiple GCN layers into one continuous layer, which is capable of leveraging fast ODE solvers and improving both performance and efficiency. All previous work aims to improve the training efficiency from the graph convolution process, whereas we innovatively challenge the necessity of time-intensive graph convolution and propose the extremely efficient post-training graph convolution framework.

\subsection{Graph Ordinary Differential Equation}

Neural ordinary-differential-equations (NODE)~\cite{Chen_node} propose a new paradigm that models continuous dynamics through the derivative of the neural network's hidden state. Motivated by this, graph ordinary-differential-equations (GDE)~\cite{poli_graph_2021} combines the concepts with GCN and directly treats the GCN layer as a continuous vector field. 
Derived from the diffusion process, continuous graph model CGNN~\cite{xhonneux2020continuous} characterizes the dynamics of node representations using a continuous message-passing layer.
Concurrently, in the continuous time data, ODE with a graph encoder \cite{luo2023hope, huang2020learning, huang2021coupled} are developed for modeling interacting dynamics. In comparison, instead of using a deep neural network to parametrize the ODE derivative, we derive the continuous graph ODE based on the discrete non-parametric graph convolution for efficient recommendation.

\section{Conclusion}

In this study, we critically challenge the conventional reliance on graph convolution in the training of graph RecSys by demonstrating that their primary benefits are realized during the testing phase. We propose the Light Post-Training Graph-ODE (LightGODE), which innovatively skips traditional resource-heavy convolution processes, and devise a novel continuous graph ordinary-differential equation model to mitigate the embedding discrepancy for optimal convolution depth. Our empirical evaluations across three real-world datasets, especially on the large-scale dataset Gowalla, show that LightGODE significantly outperforms traditional CF models in both recommendation performance and computational efficiency. This work not only questions existing training paradigms but also pinpoints potentially new research directions for efficient and large-scale graph RecSys.

\begin{acks}
This work is supported in part by NSF under grants III-2106758, and POSE-2346158
\end{acks}

\appendix

\clearpage
\section{Experimental Setup}\label{append: experiment}

\subsection{\textbf{Baselines}}
\begin{itemize}[leftmargin=18pt]
    \item \textbf{BiasMF} \cite{koren2009matrix} is a matrix factorization technique that integrates bias vectors for both users and items for enhanced prediction.
    \item \textbf{NeuMF} \cite{he2017neural} leverages deep neural networks to model the complex and non-linear interactions between users and items.
    \item \textbf{NGCF} \cite{wang2019neural} introduces GCN models with collaborative filtering to exploit the rich user-item interaction graph structure.
    \item \textbf{DGCF} \cite{wang2020disentangled} utilizes a disentangled representation learning approach to exploit distinct factors of user-item interactions. 
    \item \textbf{SimpleX} \cite{mao2021simplex} propose a novel cosine contrastive loss function to be integrated with simple collaborative filtering models.
    \item \textbf{LightGCN} \cite{he2020lightgcn} simplifies the GCN architecture for a recommendation via obviating the complex non-linear operation. 
    \item \textbf{ODECF} \cite{xu2023graph} presents a neural ODE-based model that can skip multiple GCN layers to reach the final representation. 
    \item \textbf{DirectAU} \cite{wang2022towards} explores to directly optimize the alignment and uniformity of latent representations in collaborative filtering.
    \item \textbf{GraphAU} \cite{yang2023graph} extend the alignment loss layer-wise and tailor for graph encoders for efficient graph recommendation.

\end{itemize}

\subsection{\textbf{Evaluation Metrics}}
For evaluating performance metrics, we use NDCG@K and Recall@K to ensure a fair comparison among all baseline methods in the top-K recommendation tasks. In all experiments, K is set to 20 by default unless specified. We employ the full-ranking strategy \cite{zhao2020revisiting} for all experiments, meaning that all candidate items not previously interacted by the user will be ranked during testing.

\subsection{\textbf{Implementation Details}}
Our implementation of LightGODE and all baseline models are carried out using RecBole \cite{zhao2022recbole2}. For the baseline training, we meticulously search their hyperparameters for various datasets following respective original papers to ensure a fair comparison. The batch size and the embedding size are standardized at 256 and 64, respectively. All models employ the Adam optimizer \cite{kingma2014adam}, with a learning rate set at 1e-3. To prevent overfitting, we utilize an early stopping mechanism that terminates training if there is a consistent decline in the performance metric NGCG@20 over 10 epochs. Specifically, for our method LightGODE, we tune the uniformity weight $\gamma$ within the set [0.2, 0.5, 1, 2, 5, 10, 15, 20] and search time $t$ in the range of [0.5, 0.8, 1.0, 1.2, 1.5, 1.8, 2.0, 2.2, 2.5, 3.0, 3.5, 5.0] for optimal performance. To maintain impartiality in our efficiency evaluations, each model is trained independently using a single GPU.

\section{Derivation of Continuous ODE} \label{append: ode}
One can view final representations in Equation~\ref{eq: feature} as a Riemann sum:
\begin{equation}\label{eq: Riemann sum}
\mathbf{h}^{(K)} = \sum_{k=1}^{K+1} \mathbf{A}^{(k-1)\Delta t} h_0 \Delta t,
\end{equation}
with $\Delta t = \frac{t+1}{K+1}$.  In this discrete setup, $t=K$ and thus $\Delta t = 1$ for the discrete graph convolution networks. Now let $K \rightarrow \infty $, the equation transitions to its continuous form:.
\begin{equation}\label{eq: continuous Riemann}
 \mathbf{h}(t)= \int_0^{t+1} \mathbf{{A}^s\mathbf{h}_0} \, \mathrm{d}s.
\end{equation}

Differentiating this, we have:
\begin{equation}\label{eq: diff}
\frac{\mathrm{d} \mathbf{h}(t)}{\mathrm{d} t}= \mathbf{A}^{t+1} \mathbf{h}_0.
\end{equation}

Given the challenges in computing $\mathbf{A}^{t+1}$ for non-integer values of t, it is reformulated into an ODE using the second derivative:
\begin{equation}
\frac{\mathrm{d}^2 \mathbf{h}(t)}{\mathrm{d} t^2}=\ln \mathbf{A} \mathbf{A}^{t+1} \mathbf{h}_0=\ln \mathbf{A} \frac{\mathrm{d} \mathbf{h}(t)}{\mathrm{d} t}
\end{equation}

The ODE integrates to:

\begin{equation}\label{eq: diff2}
\frac{\mathrm{d} \mathbf{h}(t)}{\mathrm{d} t}=\ln \mathbf{A} \mathbf{h}(t) + X_{const},
\end{equation}

Applying $t=0$ to Equation~\ref{eq: diff} and Equation~\ref{eq: diff2} gives:
\begin{equation}
\left.\frac{\mathrm{d} \mathbf{h}(t)}{\mathrm{d} t}\right|_{t=0}=\mathbf{A} \mathbf{h}_0=\ln \mathbf{A} \mathbf{h}_0 + X_{const}.
\end{equation}

Therefore,
\begin{equation}
X_{const} = (\mathbf{A} - \ln \mathbf{A}) \mathbf{h}_0,
\end{equation}
resulting in the ODE formulation for the graph convolution as:
\begin{equation}
\frac{\mathrm{d} \mathbf{h}(t)}{\mathrm{d} t}=\ln \mathbf{A} \mathbf{h}(t) + (\mathbf{A} - \ln \mathbf{A}) \mathbf{h}_0.
\end{equation}

\section{Analysis on the Alignment force} \label{append: align}

\begin{definition}[Perfect Alignment] A pair of observed user-item pair is perfectly aligned if $e_u = e_v$ and $(u,v) \sim P_{pos}$
\end{definition}


To simplify the derivation process, we consider the given user-item pair $(u, v)$ perfectly aligned, and the number of users is less than the number of items. The lower bound of the alignment force for the MF model is held as: 
\begin{equation}\label{eq: mf}
\begin{aligned}
\mathcal{L}_{align-mf} & = \|{e^{0}_u} - e^0_v \|^2 + \sum_{i\in N_u} \|{e^{0}_i} - e^0_u \|^2 + \sum_{j\in N_v} \|{e^{0}_v} - e^0_j \|^2 \\ 
& \geq \sum_{i\in N_u} \|{e^{0}_i} - e^0_u \|^2 + \sum_{j\in N_v} \|{e^{0}_v} - e^0_j \|^2 \\ 
& \geq \left\| \sum_{i\in N_u} ({e^{0}_i} - e^0_u) + \sum_{j\in N_v} ({e^{0}_v} - e^0_j) \right\|^2 \\ 
& \geq \left\| \sum_{i\in N_u} {e^{0}_i} - \sum_{j\in N_v} {e^{0}_j}\right\|^2, \\ 
\end{aligned}
\end{equation}
where $e^0_u$ and $e^0_v$ represents the initial embedding of user $u$ and item $v$, with $i$ and $j$ being their neighboring nodes.
Assuming the GCN model employs light convolution as in \cite{he2020lightgcn} for neighborhood aggregation, the alignment force for a 1-layer graph convolution for the user-item pair $(u, v)$ can be described as:
\begin{equation}\label{eq: gcn}
\begin{aligned}
\mathcal{L}_{align-gcn} & = \left\| {e^{1}_u} - {e^{1}_v}\right\|^2 \\
&  = \left\| \sum_{i\in N_u} \frac{e^{0}_i}{\sqrt{|N_u|} \sqrt{|N_i|}} - \sum_{j\in N_v} \frac{e^{0}_j}{\sqrt{|N_u|} \sqrt{|N_i|}} \right\|^2. \\
\end{aligned}
\end{equation}
Therefore, comparing Equation~\ref{eq: mf} and Equation~\ref{eq: gcn}, the alignment force applied on surrounding neighbors of positive pairs using graph convolution is the weighted version of the direct alignment force applied on two clusterings of nodes.

\newpage
\bibliographystyle{ACM-Reference-Format}
\bibliography{sample-base}

\end{document}